# TatBLiMP: A Benchmark of Linguistic Minimal Pairs for Tatar

**Authors.** Ilshat Saetov, Dmitry Gaynullin 

## Abstract

We introduce TatBLiMP, the first benchmark of linguistic minimal pairs for Tatar (`tt`, ISO 639-3 `tat`), a Qypchaq Turkic language written in Cyrillic. To our knowledge it is the first grammaticality evaluation for Tatar language models of any kind, since even the 101-language MultiBLiMP does not include Tatar. TatBLiMP covers 16 morphosyntactic phenomena in 1248 sentence pairs. Each pair differs by a single morpheme, one grammatical and one ungrammatical. A model passes a pair when it assigns higher probability to the grammatical member. Scoring compares probabilities the model already assigns, so the benchmark needs no text generation and no parser, and it runs on base models and on mid-training checkpoints. TatBLiMP adapts the phenomenon inventory and single-morpheme breaking operations of TurBLiMP to Tatar and adds one phenomenon specific to Tatar, bare-noun number after numerals and quantifiers. Its construction departs from template-based suites. The grammatical member of every pair is an attested sentence from Tatar literary prose. The ungrammatical member is produced by a deterministic single-morpheme perturbation with the apertium-tat transducer. Every pair is ratified by a native speaker. A plausibility principle governs construction, so the ungrammatical member is a plausible real-world error rather than an arbitrary corruption. Across from-scratch Tatar models, cross-lingual adaptations, and frontier multilingual LLMs, the benchmark tracks focused Tatar training rather than parameter scale. A 478M from-scratch model and a 125M monolingual model lead near 0.97, a 7B adaptation trails, frontier LLMs of 30–120B parameters fall to 0.80–0.92, and a lightly tuned multilingual model is weakest. A byte-normalized metric widens the gap. We close with the benchmark's main limitation. Its inherited taxonomy omits the morphophonology, vowel harmony and consonant assimilation, that is most salient to native speakers, and we sketch a native second layer that would add it.

## 1. Introduction

A minimal change to a sentence can flip it from acceptable to unacceptable for a native speaker (Chomsky 1965). Minimal-pair benchmarks turn this into a probe of a language model's grammatical knowledge. They present a grammatical sentence and a near-identical ungrammatical one and check whether the model prefers the grammatical member. BLiMP (Warstadt et al. 2020) established the format for English, and adaptations have followed for many languages, most recently TurBLiMP for Turkish (Başar et al. 2025). Tatar has had none.

Tatar, a morphologically rich agglutinative Qypchaq Turkic language written in Cyrillic, has roughly five million speakers and few digital resources. After register and quality cleaning, the clean native literary canon holds well under a billion tokens. Its speaker base is contracting, and evaluation resources for Tatar are scarce and time-sensitive. Existing Tatar model evaluation leans on corpus-fit metrics such as bits-per-byte. Those metrics measure how well a model predicts text. They do not certify that a model tells grammatical Tatar from ungrammatical Tatar.

TatBLiMP fills this gap. It probes Tatar grammatical knowledge at the level of recognition. We make the following contributions.

1. **The first grammaticality benchmark for Tatar.** 16 phenomena and 1248 native-ratified pairs, in the BLiMP tradition, adapting the TurBLiMP inventory to Tatar.
2. **A construction method for a low-resource literary language.** It uses attested literary carriers, deterministic single-morpheme perturbation with apertium-tat, native ratification of every pair, and a plausibility principle aimed at the attested degradation mode.
3. **A Tatar-specific phenomenon** beyond the TurBLiMP set, bare-noun number after numerals and quantifiers, a contrast with no counterpart in the Turkish inventory.
4. **An evaluation** across trained-Tatar models, adaptations, and frontier LLMs that shows the benchmark separates native from non-native grammatical competence, and that Tatar-specific training outperforms parameter scale.

We also state plainly what TatBLiMP does not do (§8). A high score shows recognition, which is weaker than production, and the inherited taxonomy omits Tatar morphophonology. Both bound the claims and motivate a native second layer.

## 2. Minimal-Pair Benchmarks

Minimal pairs have long been used to evaluate the linguistic abilities of language models, on phenomena such as subject–verb agreement (Linzen et al. 2016), filler-gap dependencies (Wilcox et al. 2018), and negative-polarity items (Jumelet and Hupkes 2018). Warstadt et al. (2020) scaled this into BLiMP, 67 paradigms of automatically generated English pairs. Later adaptations differ mainly in how they create pairs. Template-based generation follows BLiMP, as in CLiMP (Xiang et al. 2021), ZhoBLiMP (Liu et al. 2024), and BLiMP-NL (Suijkerbuijk et al. 2025). Another family modifies Universal Dependency trees, as in SLING (Song et al. 2022), RuBLiMP (Taktasheva et al. 2024), and the 101-language MultiBLiMP (Jumelet et al. 2026). Others extract pairs from linguistics journals (JBLiMP, Someya and Oseki 2023). These strategies trade off scale, coverage, and quality. Templates scale well but risk unnatural sentences, whereas manual and attested-text methods buy quality at the cost of scale. TurBLiMP (Başar et al. 2025) is the closest model for our work, with 16 phenomena and 1000 pairs each for a flexible-order, morphologically rich Turkic language, built with a hybrid method and native validation.

Template generation for Tatar is limited by the absence of broad-coverage generation grammars. More importantly, the failure mode we most want to detect is foreign-patterned Tatar rather than random corruption. In text this is russification, the calqued Russian-patterned Tatar of the journalistic and administrative registers. In model output it is the English structure that English-dominant models carry. TatBLiMP therefore keeps the grammatical member attested and human-vetted, and it makes the ungrammatical member a plausible error instead of an arbitrary one. MultiBLiMP does not include Tatar, and its UD-based pipeline measures agreement alone, reaching a single phenomenon even for the Qypchaq languages it does cover. The phenomenon breadth of TatBLiMP needs the hand-built route.

## 3. Tatar Morphosyntax

Tatar realizes much of its grammar through morphology instead of through function words. Case, possession, subordination, negation, evidentiality, and polarity are all suffixal, stacked in an agglutinative chain on a head-final skeleton whose canonical order is Subject-Object-Verb. Several properties make it a productive target for BLiMP-style evaluation and shape the phenomena we test.

- **Agglutinative subordination.** Relative and complement clauses form by nominalizing the embedded verb, with the participial *-ган*, the action nominal *-у*, and others, together with possessive agreement, instead of by relative pronouns or complementizers. Subordination is a morphological process more than a word-order one.
- **Izafet and possessive agreement.** Nominal modification and possession are marked by genitive and possessive concord, and the head noun carries a possessive suffix that agrees with the possessor.
- **Flexible word order with a canonical baseline.** Constituents scramble for information structure without changing core meaning, so a word-order item has to violate a grammatical constraint and not merely a pragmatic preference.

In text and everyday usage the main distortion of Tatar is russification, meaning Russian sentence logic and government under a Tatar lexicon, which saturates the journalistic and administrative registers. In AI the main distortion is English. English-dominant models process other languages through an English-centric representation (Wendler et al. 2024; Schut et al. 2025) and carry their own sentence structure into the output, so a multilingual or adapted model imposes English grammar on Tatar. For a head-final, vowel-harmonic, non-Indo-European language either foreign prior is strongly distorting. TatBLiMP is built to detect this degradation, and its ungrammatical members are, wherever possible, the errors that this Indo-European interference produces in real Tatar.

The current inventory does not cover Tatar morphophonology, its vowel harmony and suffix assimilation. Section 8 takes that up as the benchmark’s main limitation.

## 4. TatBLiMP

### 4.1 Phenomena

TatBLiMP covers 16 phenomena. Fifteen adapt TurBLiMP phenomena and their single-morpheme breaking operations to Tatar, and one, numeral–noun number, is specific to Tatar and has no TurBLiMP counterpart. One TurBLiMP phenomenon, determiners, is left out by design. A native-speaker check found no obligatory-*bir* contrast in Tatar, so it is dropped instead of left pending. In Table 1 each breaking operation turns the grammatical form into the ungrammatical one.

**Table 1. Phenomena, breaking operations, and coverage (v1.0, 1248 pairs).**

| # | Phenomenon | Breaking operation | Pairs |
|---|---|---|---|
| 1 | anaphor agreement | reflexive *үзе-* takes the wrong person or number (*үзебез* → *үзегез*) | 79 |
| 2 | argument structure, transitive | object put in the wrong lexical case (*һәркемне* → *һәркемгә*) | 78 |
| 3 | argument structure, ditransitive | direct object accusative changed to instrumental (*папкаларны* → *папкалар белән*) | 78 |
| 4 | binding (Principle B) | reflexive replaced by a pronoun in its binding domain (*үземә* → *миңа*) | 78 |
| 5 | ellipsis (backward gapping) | second-conjunct order O-S changed to S-O | 85 |
| 6 | irregular forms | aorist form altered (*күнәрләр* → *күнерләр*) | 80 |
| 7 | island effects | wh-word moved from outside a participial clause to inside it (*кайда … калдырган* → *… кайда калдырган*) | 80 |
| 8 | nominalization | factive *-ган* changed to action nominal *-у* (*онытканын* → *онытуын*) | 70 |
| 9 | NPI licensing | NPI inserted under a non-negated predicate (*сөйләгән идем* → *һич сөйләгән идем*) | 80 |
| 10 | numeral–noun number † | counted noun pluralized after a numeral or quantifier (*биш китап* → **биш китаплар*) | 85 |
| 11 | passives | agent phrase added to an impersonal passive (*…* → *… тарафыннан …*) | 83 |
| 12 | quantifiers | partitive *күбесе/күпчелеге* placed before a bare noun | 85 |
| 13 | relative clauses | | 88 |

| # | Phenomenon | Breaking operation | Pairs |
|---|---|---|---|
| | | possessive agreement on the head dropped or mismatched (*утыныбыз* → *утын*) | |
| 14 | subject–verb agreement | subject changed while the verb is left unchanged (*көтепханәләр* → *көтепханә*, verb stays plural) | 77 |
| 15 | suspended affixation | finite affix stripped from the first conjunct (*пешердем һәм …* → *пешер һәм …*) | 47 |
| 16 | word order | scrambling inside an embedded clause | 75 |
| | **total** | | **1248** |

† This phenomenon is specific to Tatar and has no TurBLiMP counterpart. In Tatar the noun after a numeral or quantifier stays singular, so *биш китап* means “five book”. Pluralizing it copies the Russian and English pattern.

## 4.2 Construction

Each item is a pair of a grammatical and an ungrammatical sentence that differ by one morpheme, labeled with its phenomenon and the morpheme that was changed. A critical region marks where the violation surfaces, for targeted scoring.

The grammatical member comes from three kinds of source. About 32% are hand-adapted TurBLiMP items, with government and lexicon adjusted to Tatar norms and false friends between Turkish and Tatar kept as deliberately hard items. About 26% are attested sentences from a curated pool of Tatar prose and poetry. About 42% are hand-composed from corpus-verified vocabulary and collocations. Unlike template suites, the grammatical member is a real, well-formed Tatar sentence rather than a synthesized frame.

The ungrammatical member is a clone of the grammatical one with exactly one changed morpheme, most often through the apertium-tat transducer. The change is a case swap, a licensed insertion, a stripped or replaced affix, an over-marked plural, or a word-order permutation. The result is always a real Tatar word form that is wrong in this context, never a nonce word. This is the plausibility principle. The bad member is a realistic error rather than an absurd string, the kind that Indo-European influence (English, Russian) produces in Tatar. A model that fails a pair therefore fails on grammar rather than on the rarity of the string.

A native speaker ratifies every pair and discards items where the bad member is accidentally grammatical or where meaning shifts. An origin gate governs what may enter at all. A phenomenon is admitted only when it ports from TurBLiMP with a native-speaker verdict, or is an independently documented calque with a source and an example.

### 4.3 Metric

The primary metric is `acc_norm`, the byte-normalized accuracy. Each member is scored on `logP / utf8_bytes`, which is tokenizer-independent and removes the advantage a model can gain from sequence length alone. The secondary metric is raw `acc`, which checks `logP(good) > logP(bad)`. Chance is 0.5 for both. Reporting both is deliberate, because agreement across the raw and the byte-normalized metric is what tells grammatical knowledge apart from a length or rarity artifact.

## 5. Experimental Setup

We evaluate models under one harness with a shared BOS prefix.

- **Trained-Tatar, from scratch.** TATlit-478M, a from-scratch Tatar literary model, and Goldfish-125M (Chang et al. 2024), a monolingual Tatar model trained on 1 GB or less in a single stage.
- **Cross-lingual adaptations.** Tweety-7B-tatar (Remy et al. 2024), trans-tokenized from Mistral, and mGPT-1.3B-tatar (Shliazhko et al. 2024), a lightly tuned multilingual base.
- **Frontier multilingual LLMs.** Gemma-4-31B, Llama-4-Scout (109B MoE, 17B active), Qwen3-32B, the only frontier model that officially documents Tatar, and gpt-oss-120b, all in bf16.

Cross-model numbers come from a single unified run on the 1248-pair build, with each model scored through its standard likelihood interface. As a cost reference, a 478M model scores all 1248 pairs in 37.2 s on one A100-80GB.

## 6. Results

Table 2 gives overall accuracy and byte-normalized accuracy. In the Tatar column, “trained” means the model was trained or tuned on Tatar, “documented” means Tatar appears in the model’s official language list, and “no” means neither.

**Table 2. Overall results, unified run (1248 pairs).**

| Model | Class | Tatar | acc | acc_norm |
|---|---|---|---|---|
| TATlit-478M | from-scratch 478M | trained | 0.975 | 0.958 |
| Goldfish-125M | external, from-scratch | trained | 0.974 | 0.958 |
| Tweety-7B | adaptation (Mistral) | trained | 0.956 | 0.915 |
| Gemma-4-31B | frontier | no | 0.924 | 0.839 |
| Llama-4-Scout | frontier (109B MoE) | no | 0.889 | 0.806 |
| Qwen3-32B | frontier | documented | 0.811 | 0.708 |
| gpt-oss-120b | frontier | no | 0.803 | 0.677 |
| mGPT-1.3B | multilingual finetune | trained | 0.736 | 0.639 |

Focused Tatar training outperforms scale. A 478M from-scratch model and a 125M monolingual model sit at the top near 0.97, a 7B adaptation is just below them at 0.96, frontier LLMs of 30–120B parameters land at 0.80–0.92, and mGPT, a small multilingual model with only light Tatar tuning, is weakest at 0.74. The ranking follows how directly a model was trained on Tatar. Frontier models bring scale without Tatar focus, and mGPT is both small and only lightly tuned, so both fall below the from-scratch and adapted Tatar models. Documented Tatar support alone is not enough either. The one frontier model that officially lists Tatar, Qwen3-32B, scores below Gemma and Llama, which do not list it, so on Tatar grammaticality scale and incidental pretraining exposure matter more than a support label. Byte-normalization widens the gap. The focused Tatar models lose only 0.02–0.04 from acc to acc_norm, whereas the frontier models and mGPT lose 0.08–0.13, for example Qwen from 0.811 to 0.708 and gpt-oss from 0.803 to 0.677, so the per-byte metric separates the classes more sharply than raw accuracy.

Per phenomenon, the trained-Tatar models are at or near ceiling across most of the inventory. The residual difficulty sits in binding, nominalization, and the subordination phenomena of island effects and relative clauses, while the rest of the inventory is saturated.

Islands are near chance only for the frontier. The single lowest per-phenomenon score in the whole evaluation is Qwen3-32B on island effects at 0.512, which is chance, alongside its lows on irregular forms (0.700) and subject–verb agreement (0.701). The trained-Tatar models are far from chance on the same items. The island items actually manipulate the position of a wh-word inside or outside a participial clause, which is a different thing from an English-style extraction island. Section 8 returns to this.

## 7. Discussion

TatBLiMP ranks from-scratch and monolingual Tatar models above adaptations and far above frontier multilingual LLMs, and it does so most sharply under the byte-normalized metric that removes length effects. Because it runs on a likelihood comparison with no generation and no parser, it can score base models and mid-training checkpoints in seconds, which makes it usable as an in-training signal for tokenizer and data ablations and not only as a final report card.

The focused Tatar models cluster near 1.0 on most phenomena, and the discriminating signal concentrates in nominalization, binding, and the subordination phenomena of islands and relative clauses. Tatar grammaticality, as this inventory defines it, is largely solved by a model with focused Tatar training, and the residual difficulty sits in morphological subordination. That is a real finding, and it is also a warning about coverage, which the next section makes explicit.

## 8. Limitations and Future Work

**Recognition and production.** A high TatBLiMP score shows that a model detects violations. It does not show that the model generates native Tatar. Verb-finality, SOV adherence, and foreign-structure drift on generated text need an instruction-tuned generator and a production harness, which we defer to separate work.

**Register.** Carriers are literary prose, which may under-represent colloquial usage. A colloquial layer is future work.

**No separate human-acceptability study yet.** A native speaker ratifies every pair during construction, but we have not collected an independent set of graded human acceptability judgments in the style of TurBLiMP's human baseline. That baseline would let us report a human and model gap per phenomenon, and it is planned.

**The inherited taxonomy omits Tatar morphophonology.** This is the benchmark's most important limitation, and it is structural. The inventory descends from English BLiMP through TurBLiMP. The phenomenon set was chosen, twice removed, for what English and Turkish syntax foreground, among them binding, NPIs, islands, filler-gap, and anaphor agreement. It was never built from Tatar morphology.

The skew shows up in our own numbers. Most transplanted phenomena are saturated for any competent Tatar model (§6), and the one place the frontier collapses to chance, islands, turns out to test wh-placement in a participial clause instead of a genuine extraction island, a contrast that Tatar barely has. Meanwhile the layer a native speaker hears first, vowel harmony and suffix assimilation, is absent from the benchmark. That absence follows directly from the construction rule. The origin gate admits a phenomenon only when it ports from TurBLiMP or is a documented calque, and the plausibility principle requires the ungrammatical member to be a real word. A harmony violation such as *китаплар* → **китаплəр* is by definition a non-word, so it passes neither gate. English BLiMP could ban non-words at no cost because English has almost no morphophonology. The same ban on Tatar removes a large and central part of the grammar.

The fix is to separate the two things the never-nonce rule conflates. The first is the phenomenon set, which is genuinely anglocentric and should be rebuilt from Tatar morphology. The relevant axes are ones English cannot lend a taxonomy for, among them vowel harmony in backness and rounding, suffix consonant assimilation (*-да/-та*, *-га/-ка*, *-дан/-тан*), affix ordering in the agglutinative chain, izafet with its three possessive types, evidentiality (*-ган* against *-ды*), postposition case government, and converbs. The second thing is measurement hygiene, which is legitimate and language-independent. The rule exists so that a model cannot win a pair merely because the bad string is rare or out-of-vocabulary, and this confound is real, since a low log-probability can come from rarity instead of from grammatical knowledge. For harmony the confound is weak. *китаплар* and *китаплəр* are both orthographically well-formed, both pronounceable, and they differ by a single vowel. A model that confidently ranks *китаплəр* lower is showing knowledge of harmony. It is not reacting to an unfamiliar string. Harmony should therefore be controlled rather than banned. The controls are a minimal one-segment flip, matched stem frequency across the pair, and reporting on both the raw and the byte-normalized metric, the same instrument that already separates frontier from native models in §6.

The remedy keeps TatBLiMP and adds a second, orthogonal layer in which the never-nonce rule is lifted for morphophonology, where a violation is a non-word by definition and that is exactly the phenomenon. The layer would have two families. Morphophonological pairs would cover vowel harmony and suffix assimilation, where the bad member is an intentional non-word under the controls above. Agglutinative pairs would cover affix order, izafet, postposition government, and converbs, where the bad member is sometimes a real word and sometimes a non-word. The current syntactic layer asks whether a model recognizes native structure in tasks that have an English analogue. The native layer would ask whether it holds the structure that has no English analogue, the part a Tatar speaker notices first. Together they measure coverage across the language's real axes, not only fidelity of execution on a borrowed inventory.

**Skewed difficulty profile.** The balanced design that keeps the phenomena comparable also skews the difficulty distribution. Each phenomenon contributes a similar number of pairs, most of them sit at ceiling for a competent Tatar model (§6), and the aggregate score is dominated by items that no trained model misses. In psychometric terms the suite is short of hard and mid-difficulty items, so its discriminating power rests on a few phenomena. Reweighting phenomena by discriminating power would sharpen the ranking, but the weights would be fitted to the current model pool and would not transfer to other model classes, so raw accuracy stays the metric and the per-phenomenon breakdown carries the structure. Rebalancing item by item is impractical within a fixed per-phenomenon budget and would still draw on the borrowed inventory. The proper remedy is the native second layer above, a separate benchmark built from Tatar's own structure through a native harness, and that work has begun.

## 9. Ethics and Licensing

TatBLiMP is released under CC BY-NC 4.0. It is free for non-commercial use, sharing, and adaptation with attribution, and items adapted from TurBLiMP derive from CC BY 4.0 material and are attributed accordingly. The benchmark measures grammatical competence only. It is not a measure of a model's overall quality, knowledge, or generation ability.

## Citation

```
@misc{tatblimp2026,
  title  = {TatBLiMP: A Benchmark of Linguistic Minimal Pairs for Tatar},
  author = {Ilshat Saetov and Dmitry Gaynullin},
  year   = {2026},
  url    = {https://huggingface.co/datasets/ilchats/TatBLiMP}
}
```

Built on BLiMP (Warstadt et al. 2020), TurBLiMP (Başar et al. 2025), MultiBLiMP (Jumelet et al. 2026), and the apertium-tat transducer (Washington et al. 2014).

---

## References

Ezgi Başar, Francesca Padovani, Jaap Jumelet, and Arianna Bisazza. 2025. TurBLiMP: A Turkish Benchmark of Linguistic Minimal Pairs. In Proceedings of the 2025 Conference on Empirical Methods in Natural Language Processing, pages 16495–16510. https://aclanthology.org/2025.emnlp-main.834/

Tyler A. Chang, Catherine Arnett, Zhuowen Tu, and Benjamin K. Bergen. 2024. Goldfish: Monolingual Language Models for 350 Languages. arXiv:2408.10441.

Noam Chomsky. 1965. Aspects of the Theory of Syntax. MIT Press, Cambridge, MA.

Jaap Jumelet and Dieuwke Hupkes. 2018. Do Language Models Understand Anything? On the Ability of LSTMs to Understand Negative Polarity Items. In Proceedings of the 2018 EMNLP Workshop BlackboxNLP, pages 222–231. https://aclanthology.org/W18-5424/

Jaap Jumelet, Leonie Weissweiler, Joakim Nivre, and Arianna Bisazza. 2026. MultiBLiMP 1.0: A Massively Multilingual Benchmark of Linguistic Minimal Pairs. Transactions of the Association for Computational Linguistics, 14:193–216. https://aclanthology.org/2026.tacl-1.10/

Tal Linzen, Emmanuel Dupoux, and Yoav Goldberg. 2016. Assessing the Ability of LSTMs to Learn Syntax-Sensitive Dependencies. Transactions of the Association for Computational Linguistics, 4:521–535. https://aclanthology.org/Q16-1037/

Yikang Liu, Yeting Shen, Hongao Zhu, Lilong Xu, Zhiheng Qian, Siyuan Song, Kejia Zhang, Jialong Tang, Pei Zhang, Baosong Yang, Rui Wang, and Hai Hu. 2024. ZhoBLiMP: A Systematic Assessment of Language Models with Linguistic Minimal Pairs in Chinese. arXiv:2411.06096.

François Remy, Pieter Delobelle, Hayastan Avetisyan, Alfiya Khabibullina, Miryam de Lhoneux, and Thomas Demeester. 2024. Trans-Tokenization and Cross-lingual Vocabulary Transfers: Language Adaptation of LLMs for Low-Resource NLP. In First Conference on Language Modeling (COLM 2024). arXiv:2408.04303.

Lisa Schut, Yarin Gal, and Sebastian Farquhar. 2025. Do Multilingual LLMs Think In English? arXiv:2502.15603.

Oleh Shliazhko, Alena Fenogenova, Maria Tikhonova, Anastasia Kozlova, Vladislav Mikhailov, and Tatiana Shavrina. 2024. mGPT: Few-Shot Learners Go Multilingual. Transactions of the Association for Computational Linguistics, 12:58–79. https://aclanthology.org/2024.tacl-1.4/

Taiga Someya and Yohei Oseki. 2023. JBLiMP: Japanese Benchmark of Linguistic Minimal Pairs. In Findings of the Association for Computational Linguistics: EACL 2023, pages 1581–1594. https://aclanthology.org/2023.findings-eacl.117/

Yixiao Song, Kalpesh Krishna, Rajesh Bhatt, and Mohit Iyyer. 2022. SLING: Sino Linguistic Evaluation of Large Language Models. In Proceedings of the 2022 Conference on Empirical Methods in Natural Language Processing, pages 4606–4634. https://aclanthology.org/2022.emnlp-main.305/

Michelle Suijkerbuijk, Zoë Prins, Marianne de Heer Kloots, Willem Zuidema, and Stefan L. Frank. 2025. BLiMP-NL: A Corpus of Dutch Minimal Pairs and Acceptability Judgments for Language Model Evaluation. Computational Linguistics, 51(4):1267–1301. https://aclanthology.org/2025.cl-4.6/

Ekaterina Taktasheva, Maxim Bazhukov, Kirill Koncha, Alena Fenogenova, Ekaterina Artemova, and Vladislav Mikhailov. 2024. RuBLiMP: Russian Benchmark of Linguistic Minimal Pairs. In Proceedings of the 2024 Conference on Empirical Methods in Natural Language Processing, pages 9268–9299. https://aclanthology.org/2024.emnlp-main.522/

Alex Warstadt, Alicia Parrish, Haokun Liu, Anhad Mohananey, Wei Peng, Sheng-Fu Wang, and Samuel R. Bowman. 2020. BLiMP: The Benchmark of Linguistic Minimal Pairs for English. Transactions of the Association for Computational Linguistics, 8:377–392. https://aclanthology.org/2020.tacl-1.25/

Jonathan Washington, Ilnar Salimzyanov, and Francis Tyers. 2014. Finite-state morphological transducers for three Kypchak languages. In Proceedings of the Ninth International Conference on Language Resources and Evaluation (LREC’14), pages 3378–3385. https://aclanthology.org/L14-1143/

Chris Wendler, Veniamin Veselovsky, Giovanni Monea, and Robert West. 2024. Do Llamas Work in English? On the Latent Language of Multilingual Transformers. In Proceedings of the 62nd Annual Meeting of the Association for Computational Linguistics (Volume 1: Long Papers), pages 15366–15394. https://aclanthology.org/2024.acl-long.820/

Ethan Wilcox, Roger Levy, Takashi Morita, and Richard Futrell. 2018. What do RNN Language Models Learn about Filler-Gap Dependencies? In Proceedings of the 2018 EMNLP Workshop BlackboxNLP, pages 211–221. https://aclanthology.org/W18-5423/

Beilei Xiang, Changbing Yang, Yu Li, Alex Warstadt, and Katharina Kann. 2021. CLiMP: A Benchmark for Chinese Language Model Evaluation. In Proceedings of the 16th Conference of the European Chapter of the Association for Computational Linguistics: Main Volume, pages 2784–2790. https://aclanthology.org/2021.eacl-main.242/